\documentclass{article} 
\usepackage{iclr2021_conference,times}

\usepackage{hyperref}
\usepackage{url}
\usepackage[percent]{overpic}
\usepackage{paralist}

\usepackage{natbib} 

\usepackage{mathtools} 
\usepackage{siunitx} 
\usepackage{booktabs} 
\usepackage{tikz} 
\usepackage{todonotes}

\usepackage{bm}
\usepackage{amsmath}
\usepackage{amssymb}
\usepackage{makeidx}
\usepackage{graphicx}
\usepackage{siunitx}
\usepackage{color}
\usepackage{colortbl}
\usepackage{gensymb}
\usepackage{subcaption}
\usepackage{multirow}
\usepackage{paralist}

\usepackage{diagbox}
\usepackage{float}

\usepackage{todonotes}

\newcommand{\B}[1]{\mathbf{#1}}

\usepackage[ruled, linesnumbered]{algorithm2e}

\graphicspath{ {./figures/} }

\title{Spliced Binned-Pareto Distribution for robust modeling of heavy-tailed time series}

\author{Elena Ehrlich \thanks{Fran\c{c}ois-Xavier Aubet and Elena Ehrlich contributed equally to this work.} 
  \\
AWS ProServe \\
Miami, FL, USA \\
\texttt{eeehrlic@amazon.com} \\
\And
Laurent Callot \\
Amazon Research \\
Seattle, WA, USA \\
\texttt{lcallot@amazon.com} \\
\And
Fran\c{c}ois-Xavier Aubet$\;^*$ \\
Amazon Research \\
Vienna, Austria \\
\texttt{aubetf@amazon.com} \\
}

\iclrfinalcopy 
\begin{document}

\maketitle

\begin{abstract}
This work proposes a novel method to robustly and accurately model time series with heavy-tailed noise, in non-stationary scenarios.
%
In many practical application time series have  heavy-tailed noise that significantly impacts the performance of classical forecasting models; in particular, accurately modeling a distribution over extreme events is crucial to performing accurate time series anomaly detection.
%
%
We propose a Spliced Binned-Pareto distribution which is both robust to extreme observations and allows accurate modeling of the full distribution. 
Our method allows the capture of time dependencies in the higher order moments of the distribution such as the tail heaviness.
We compare the robustness and the accuracy of the tail estimation of our method to other state of the art methods on Twitter mentions count time series. 


\end{abstract}

\section{Introduction} \label{sec:introduction}

In many real world applications, time series can have heavy-tailed distributions, examples include: (i) 
financial series where speculators who bet high, bet very high~\citep{bradley2003financial}, (ii) server metrics like requests-per-second to plan compute resource scaling, (iii) extreme rainfall for flood prediction~\citep{bezak2014comparison}.
Extreme events are known to impede estimation of the predictive density $p(\B{x}_{t+1} | \B{x}_{1:t})$ for a time series $\B{x}_{1:T}$, $\B{x}_t\!\!\in\!\!\mathcal{X}\!\!\subset\!\!\mathbb{R}$, $t,\!T\!\!\in\!\!\mathbb{N}_1$, under classical methods.
This problem has been extensively studied in the case of independently and identically distributed (iid) data~\citep{beirlant2006statistics}, and has recently gained attention in the context of time series~\citep{kulik2020heavy,davison2015statistics}.
Accurate estimation of distribution tails can be important in forecasting settings and is crucial in time series anomaly detection, where how likely the extreme events are dictates whether an alarm should be raised.

Robust and accurate density estimation is challenging in time series with heavy-tailed noise for two main reasons:
first, extreme events have a strong impact on the estimation of the base distribution;
second, one has to account for time-varying components of the time series in order to identify the tails before beginning to fit the tails themselves.
While some methods have been proposed for tail estimation, they assume little to no time-varying components~\cite{siffer2017anomaly}.
To obtain a forecast robust to extreme values and adjustable to many shapes of distributions, discrete binned distributions can be used~\cite{rabanser2020effectiveness}, these are parametrised by a Neural Network (NN), which have been shown capable of capturing complex long-term time dependencies in forecasting~\citep{benidis2020neural}.
We address these two challenges by combining a binned distribution with Generalised Pareto  distributions for the tails, all three distributions parameterised by a single NN, allowing us to jointly model time dependencies in the base distribution and in the tails.

%
In particular, we make the following contributions:
\begin{enumerate}
	\item We propose a combination of a binned distribution, for a robust estimate of the base distribution, with Generalised Pareto distributions, for an accurate estimate of the tails.
	\item Our method allows for asymmetric tails with time-varying heaviness and scale.
	\item We show empirically on a real world example that our approach allows for a better estimate of the tail distribution than previous methods, while also providing an accurate estimate of the base distribution.
\end{enumerate}

\section{Background \& Related Work} \label{sec:background}

\paragraph{Extreme Value Theory (EVT)}
%
A classic result from extreme value theory (EVT) states that the distribution of extreme values is almost independent of the base distribution of the data~\citep{fisher1928limiting}. As a consequence, it has been proposed to estimate only the tail of the distribution by considering only the peaks above threshold $\tau\in\mathcal{X}$, the assumed upper bound of the base distribution.
Let $\bar{F}_{\tau}(x)=P(X>x+\tau|X>\tau)$
;  by the second theorem of EVT~\citep{balkema1974residual,pickands1975statistical}:
\begin{align}
  \bar{F}_{\tau}(x) &\sim \mathrm{GPD}(\xi,\beta) \label{eq:pickands}
\end{align}
where $\mathrm{GPD}$ is a Generalised Pareto Distribution with shape $\xi\in\mathbb{R}$ and scale $\beta\in\mathbb{R}^+$.
Therefore the quantile value $z_q: P(X>z_q)< q$ for quantile level $q\in[0,1]$ can be calculated
\begin{align}
  z_q &\approx \tau + \frac{\hat{\beta}}{\hat{\xi}} \left( \left(\frac{qT}{N_{\tau}}\right)^{-\hat{\xi}} -1 \right) \label{eq:zq}
\end{align}
%
%
upon solving for $\hat{\xi},\hat{\beta}$ 
and where $N_{\tau}$ is the number of peaks over initial threshold ${\tau}$ out of $T$ observations. This method of computing the quantile of extreme values is known as Peaks-Over-Threshold (POT)~\citep{LEADBETTER1991357}.
Note that Eq~\ref{eq:pickands} only holds as $\tau$ tends to the true upper bound of the base distribution. In practice $\tau$ is not known, and has to be determined.
This introduces a trade-off between the variance incurred from too few excesses above an over-estimated threshold and the bias incurred from non-tail observations introduced into the GPD fit.

\paragraph{Streaming POT and Drift SPOT}
\citet{siffer2017anomaly} extended the result of POT to the streaming setting and for incorporation of online anomaly detection in stationary and concept-drift time series, known respectively as Streaming POT (SPOT) and Drift SPOT (DSPOT).
They aim to accurately estimate a high quantile $z_q$ above which they consider incoming points as anomalous.
Using a fixed threshold $\tau$, they use POT to initialise $z_q$ from the first $T$ observations. The method then processes data arrivals of $X_t<\tau$ as `normal', $X_t>z_q$ as anomalous, and $\tau\leq X_t\leq z_q$ as non-anomalous `real' peaks that trigger a recalculation of Eq~\ref{eq:pickands}-\ref{eq:zq} to update $z_q$.

Slightly relaxing the stationarity restriction of SPOT, DSPOT subtracts a sliding window average of `normal' points before applying SPOT. Though adaptable to slow drifts in the distribution mean, DSPOT remains limited in three ways:
\begin{inparaenum}[1)]
\item it does not model complex time-dependencies in the base distribution,
\item it does not model the base distribution and is therefore not suitable for forecasting,
\item it does not allow for a time-varying, feature dependent parameterisation of the tails.
\end{inparaenum}


\paragraph{NN extensions of SPOT}
To further improve DSPOT's relaxation of the stationarity constraint,
\cite{davis2019lstm} proposed point forecasting by training an RNN to minimise the Mean Squared Error (MSE), and then using SPOT on the prediction residuals to detect anomalies.
%
The solution is limited in three ways: 1) the MSE loss function is sensitive to extreme events, 
2) no allowance is made for time dependencies in moments of the distribution beyond the mean, be it the variance or the tail heaviness, 3) the use of residuals prohibits modelling asymmetric tails.

\section{Spliced Binned-Pareto distribution} \label{sec:method}

We propose the Spliced Binned-Pareto (SBP) distribution which uses a flexible binned distribution to model the base of the distribution and two $\mathrm{GPD}$s to model the tails.

\begin{figure}[h]
	\begin{subfigure}{0.5\columnwidth}
		\centering
		\includegraphics[width=0.99\columnwidth]{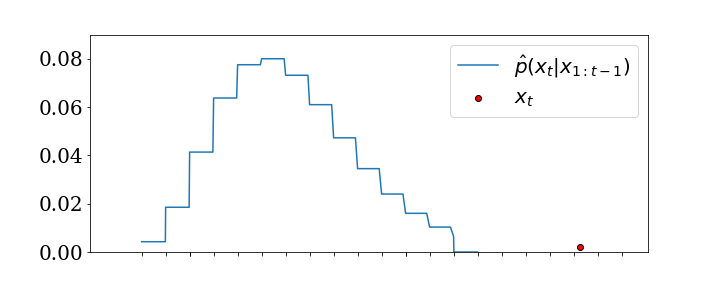}
		\caption{Predictive binned distribution.}
		\label{fig:visual1}
	\end{subfigure}%
	\centering	
	\begin{subfigure}{0.5\columnwidth}
		\centering
		\includegraphics[width=0.99\columnwidth]{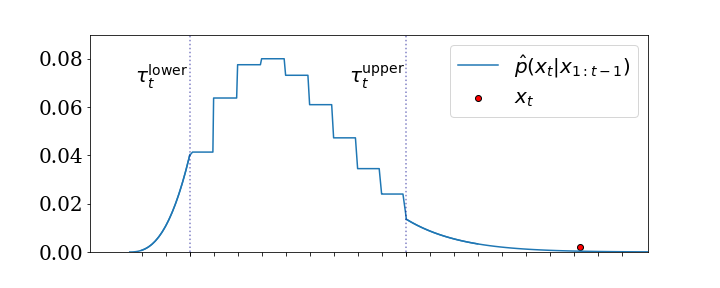}
		\caption{Predictive Spliced Binned-Pareto distribution.}
		\label{fig:visual2}
	\end{subfigure}	
	\centering		
	\caption{Illustration of the parametric tail accounting for an extreme event.}
	\label{fig:visual}
\end{figure}

We model the base of the predictive distribution with a discrete binned distribution to make it robust to extreme values and adaptable to the variety of real work distributions. As described by \citet{rabanser2020effectiveness}, we discretise the real axis between two points into $n$ bins.
A NN is trained to predict the probability of the next point falling in each of these bins, as shown in Figure \ref{fig:visual1}. 
This gives a distribution robust to extreme values at training time because it is now a classification problem, the log-likelihood is not affected by the distance between the predicted mean the observed point, as would be the case when using a Gaussian or Student's-t distribution for example.



We enhance the binned distribution to have an accurate parametric estimate of the tails by replacing the tails of the binned with $\mathrm{GPD}$s.
From the binned distribution, we deliminate the base distribution from its upper and lower tails according to user-defined quantiles $q$ and $1-q$ respectively  (e.g.\ $q = 0.05$).
At time $t$ we obtain $\tau_t^{\text{lower}} = z_{1-q}{(t)}$ from the predictive binned distribution and replace the cumulative distribution function (cdf) of the bins below this quantile with the weighted cdf of the lower $\mathrm{GPD}$. Analogously we obtain $\tau_t^{\text{upper}} = z_{q}{(t)}$ and replace the cdf above with the weighted upper $\mathrm{GPD}$.
Using this procedure we obtain a valid probability distribution, integrating to 1, with support $(-\infty , \infty) $ (illustration in Figure~\ref{fig:visual2}).

The binned distribution and the $\mathrm{GPD}$s are parameterised by a single NN taking as input $\B{x}_{1:t-1}$, the past points, and outputting $n +4$ parameters: each of the $n$ bin probabilities as well as $\xi$ and $\beta$ for each of the $\mathrm{GPD}$.
While our approach is fully general and works with most NN architectures, we opt to use a Temporal Convolution Network (TCN)~\citep{bai2018empirical}. We expect our results to carry over to other architectures.

In addition to being robust to extreme values, it also results in a robust estimate of the tails of the distributions as it can model changes in the tails over time.
While predecessor methods 
 use a fixed threshold $\tau$ to delimit the tails, by modeling the base distribution we obtain a time-varying threshold.
Furthermore, training a single NN parameterising all distributions to maximise the log probability of the observed time step under the binned and $\mathrm{GPD}$ distributions, results in an prediction that accounts for temporal variation in all moments of the distribution: the mean and variance as well as tail heaviness and scale. This includes asymmetric tails.

\section{Experiments} \label{sec:experiments}

We make our implementation available in GitHub repository \url{https://github.com/awslabs/gluon-ts/blob/dev/src/gluonts/torch/distributions/spliced_binned_pareto.py}. 
We compare our approach to the different methods presented in Related Work: SPOT, DSPOT, as well as SPOT on NN prediction residuals. We implemented \citet{davis2019lstm}  using a TCN to restrict ourselves to differences in the extreme value handling; we refer to it as TCN-SPOT.
%
Following \citep{siffer2017anomaly}, for each of SPOT, DSPOT and TCN-SPOT, we set threshold $\tau$ at $z_{0.95}$ of the training data for the lower tail and at $z_{0.05}$ for the upper tail.
And we set the thresholds of the SBP with $q = 0.05$, replacing the lower and upper 5\% of the distribution with $\mathrm{GPD}$s.

\paragraph{Evaluation metric}
We evaluate the accuracy of the density estimation of each of the method using Probability-Probability plots (PP-plots)~\cite{michael1983stabilized}.
For a given quantile level $q$, we compute $ y_q$ the fraction of points that fell below the given quantile $z_q{(t)} $ of their corresponding predictive distribution:
\begin{align}
  y_q = \frac{\sum_{t=2}^{T} \mathbb{I}[ \B{x}_t < z_{1-q}{(t)} ] }{T}, \hspace{40pt}   z_q{(t)} : p\left( \B{x}_{t} > z_q{(t)} \middle| \B{x}_{1:t-1} \right)< q
\end{align}
To obtain a quantitative score, we measure how good the tail estimate is by computing the Mean Absolute Error (MAE) between  $y_q $ and $q $ for all measured quantiles $q$.

\subsection{Evaluation on synthetic data}

\begin{figure}[h]
	\begin{subfigure}{0.4\columnwidth}
		\centering
		\includegraphics[width=0.99\columnwidth]{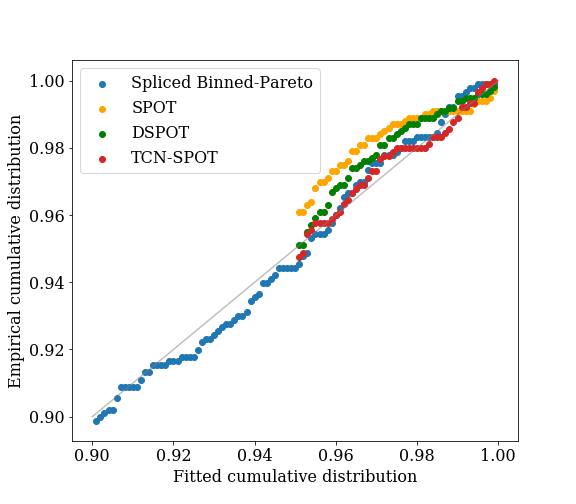}
		\caption{PP-plot on the sinwave time series for $\rm{cdf}\in[0.9, 1.0)$.}
		\label{fig:synth1}
	\end{subfigure}%
	\centering	
	\begin{subfigure}{0.5\columnwidth}
				\renewcommand{\arraystretch}{1.1}
				\begin{tabular}{lccccccccccccc}
					\toprule
					Model & Synthetic data & Real data \\ 
					\midrule
					SPOT  & $9.0\times10^{-3}$ & $9.61\times10^{-3}$  \\
					DSPOT  & $5.65\times10^{-3}$&  $1.08\times10^{-2}$  \\
					TCN-SPOT  & $\bm{2.06\times10^{-3}}$ &  $8.77\times10^{-3}$  \\
					SBP (ours) &  $3.17\times10^{-3}$ & $\bm{4.24\times10^{-3}}$ \\
					\bottomrule
				\end{tabular}
			\caption{Mean absolute error between the empirical quantiles and the predicted quantiles, on the synthetic and real datasets. (lower error is better)}
		\label{table:MAE_comparison}
	\end{subfigure}	
	\centering		
	\caption{Evaluation on synthetic time series and comparison of the MAE.}
	\label{fig:synth}
\end{figure}

To  compare the different methods on simple time dependencies, we generated sine waves and added iid Student's-t and heavy-tailed noises for synthetic data. 
Figure \ref{fig:synth1} shows the PP-plot for each comparative method.
We observe that our method provides an accurate estimate of the density both of the top of the base distribution, $\textrm{cdf} \in [0.90 , 0.95]$, and of the tail of the distribution, $\textrm{cdf} \in [0.95 , 1.0)$, and does so without discontinuity at $\tau_t = z_{0.05}{(t)}$.
Further, while one could have feared that parametrising the $\mathrm{GPD}$ using a NN 
could have been noisy or less robust, in fact our method obtains a better estimate of the tail than SPOT or DSPOT.
\vspace{-10pt}
%
\vspace{-10pt}
\subsection{Evaluation on a real world scenario}
%



\begin{figure}[h]
	\begin{subfigure}{0.66\columnwidth}
		\centering
                \includegraphics[width=0.99\columnwidth]{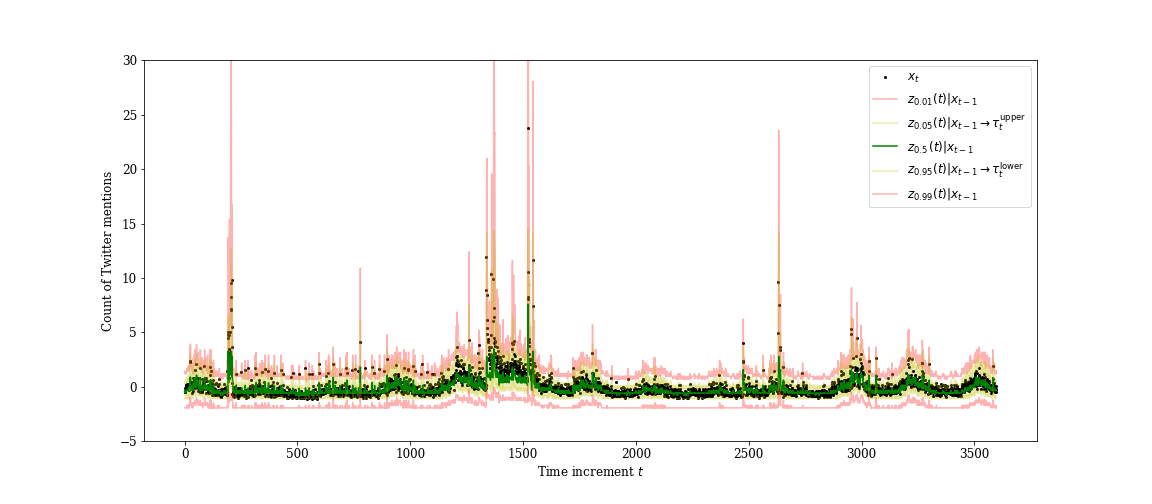}
		\caption{Spliced Binned-Pareto fit on the test set of real time series.}
		\label{fig:real1}
	\end{subfigure}%
	\centering	
	\begin{subfigure}{0.33\columnwidth}
		\centering
                \includegraphics[width=0.99\columnwidth]{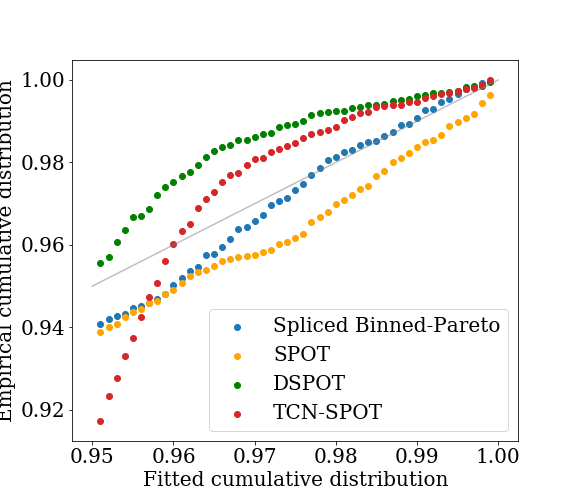}
		\caption{PP-plot of the upper tail.}
		\label{fig:real2}
	\end{subfigure}	
	\centering		
	\caption{Time series of the stock ticker's Twitter mentions per 5-minute interval}
	\label{fig:real}
\end{figure}

Retailers rely upon demand forecasting to inform inventory ordering.
Recently, the reach of social media platforms has meant that social trends take off faster than the forecast can anticipate. While the first extreme demand spike is not predictable, it is still important to reliably quantify its likelihood.
To illustrate this point, we look at 
the count of Twitter mentions that a stock ticker symbol of interest receives per 5-minute interval~(data source: Numenta~\cite{ahmad2017unsupervised}). This series exhibits seasonalities, heteroscedasticity, and extreme realizations; see Figure~\ref{fig:real1}.


Figure \ref{fig:real2} shows the PP-plots on the real time series, focusing on the tail estimate for $\rm{cdf}~\!\!\in~\!\![0.95, 1.0)$. We observe that our method obtains a significantly better tail estimate than the comparison partners. 
%
Table \ref{table:MAE_comparison} shows that, while in the setting with simple time dependencies and symmetric tail heaviness our method performs comparably to other method, in the real scenario the advantage of our method's robustness becomes clear.


%
%
%

\vspace{-10pt}
\section{Discussion}
This work presents the Spliced Binned-Pareto Distribution distribution which, combined with a TCN, allows robust and accurate estimation of the predictive density in the presence of extreme events.
%
%
The bias variance trade-off inherent to setting $\tau$ in SPOT is less present in our method as the thresholds are time-varying; however our method still requires a user-defined quantile to deliminate each tail. We want to investigate approaches to learn the tail quantile from training data to further reduce the domain knowledge needed to use the method.

\bibliography{bibliography}
\bibliographystyle{iclr2021_conference}

\appendix



\end{document}